\documentclass{article}

 \usepackage[main, final]{neurips_2026}

\usepackage[utf8]{inputenc} 
\usepackage[T1]{fontenc}    
\usepackage{hyperref}       
\usepackage{url}            
\usepackage{booktabs}       
\usepackage{amsfonts}       
\usepackage{nicefrac}       
\usepackage{microtype}      
\usepackage{xcolor}         
\usepackage{amsmath}
\usepackage{graphicx} 
\usepackage{multirow}
\usepackage{wrapfig}

\usepackage{bm}
\usepackage{tcolorbox}
\tcbuselibrary{breakable}
\usepackage{enumitem}
\usepackage{amssymb}
\usepackage{algorithm}
\usepackage{algpseudocode}

\title{Escaping Reasoning Basin Collapse with History-Biased Search}

\author{%
  Lu Cheng \\
  Department of Computer Science and Engineering\\
  Penn State University\\
  University Park, PA 10681 \\
  \texttt{lqc5822@psu.edu} \\
}

\begin{document}

\maketitle

\begin{abstract}
Inference-time search with large language models (LLMs) often concentrates on a small set of structurally or semantically similar trajectories, leaving alternative reasoning strategies underexplored---a failure mode we call \textit{reasoning basin collapse}. We introduce \textsc{BASIN}, a training-free, history-biased search method that groups reasoning states into basins and accumulates a revisit penalty on repeatedly selected basins, reallocating a fixed inference budget toward underexplored reasoning strategies. Under matched inference budgets, \textsc{BASIN} improves over Tree of Thoughts (ToT) by up to $+22$pp on Game of 24 and $+6.7$pp on MuSR. Because indiscriminate diversification can over-explore once search has found a promising basin, we further introduce \textsc{QA-BASIN}, a quality-aware variant that weakens the revisit penalty for high-quality basins and yields more robust gains. To characterize when basin-aware search helps, we introduce the \emph{redundancy gap} $\Delta$, which measures the difference in search concentration between correct and incorrect predictions: standard ToT often operates near $\Delta \approx 0$, whereas \textsc{BASIN} consistently shifts $\Delta$ positive. Together, these results identify reasoning basin collapse as a failure mode of inference-time search and show that history-dependent bias provides a simple, training-free mechanism for escaping redundant reasoning under fixed compute. Code is available at \url{https://github.com/GitHubLuCheng/basin}.
\end{abstract}

\section{Introduction}

When faced with a difficult reasoning problem, a careful human solver rarely
repeats the same line of argument indefinitely---instead, they try genuinely
different explanations before producing more variants of the same one. Yet
this discipline is not built into most LLM inference-time search procedures.
Recent work improves LLM reasoning by scaling inference-time compute,
generating multiple candidate trajectories and selecting or aggregating among
them~\citep{wei2022chain,wang2023selfconsistency,yao2023tree,besta2024graph,hao2023reasoning,zhou2024language,ding2025dynamic}.
More trajectories, however, do not necessarily imply more distinct reasoning
strategies. Standard search is largely unaware of which candidates revisit an
already explored strategy and can therefore spend substantial inference budget
producing variations of the same underlying approach. We call this failure mode
\textbf{reasoning basin collapse}.

\begin{wrapfigure}{r}{0.3\textwidth}
\centering
\vspace{-1em}
\includegraphics[width=\linewidth]{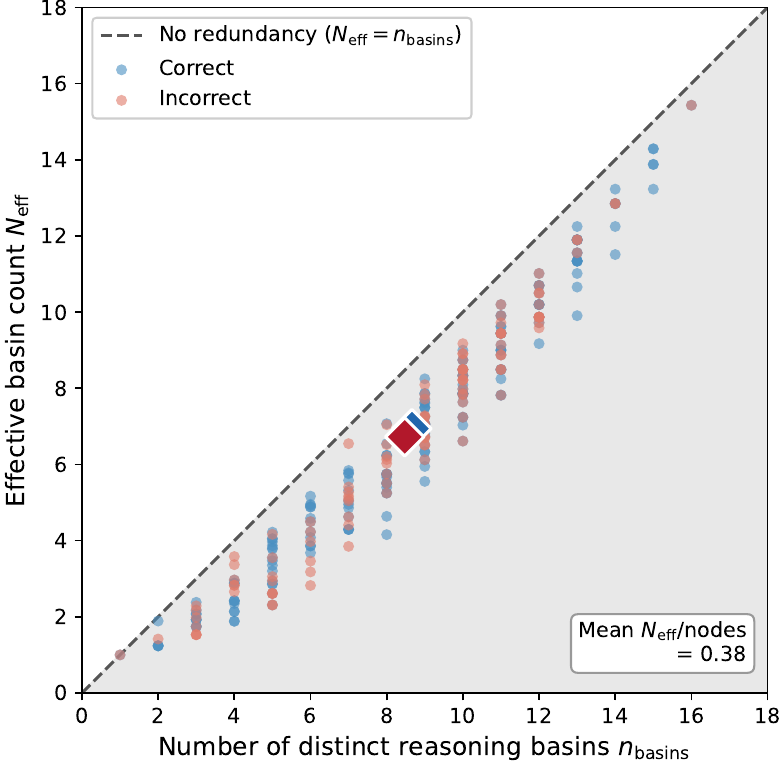}
\vspace{-1em}
\caption{
\textbf{Reasoning basin collapse in ToT.}
Each point is one MuSR problem ($n{=}300$, \texttt{gpt-oss-120b}).
The $y$-axis shows $N_{\mathrm{eff}}$, the effective number of basins explored after accounting for visit concentration.
All points fall below the diagonal (shaded), where
$N_{\mathrm{eff}} < n_{\mathrm{basins}}$, indicating that visits
concentrate on previously explored basins.
}
\label{fig:basin_collapse}
\vspace{-2em}
\end{wrapfigure}

Figure~\ref{fig:basin_collapse} illustrates this phenomenon on
MuSR~\citep{sprague2024musr}, a benchmark requiring multi-step reasoning. Under
standard Tree of Thoughts (ToT), the effective basin count
$N_{\mathrm{eff}}$ (formally defined in Sec.~\ref{sec:experimental_setup})
falls well below the number of generated states: Although ToT generates many reasoning states, its effective basin count is only 38\% of the number of generated nodes on average, indicating that much of the search budget is spent revisiting a small set of strategies. Moreover, this
collapse is \emph{indiscriminate}: incorrect searches can repeatedly revisit
wrong strategies just as readily as correct searches can concentrate on useful
ones. Search concentration alone therefore provides little indication of
whether the explored basin is actually productive.

To formalize this behavior, we define a \textbf{reasoning basin} as an
equivalence class of reasoning states that pursue the same underlying strategy.
For tasks with explicit structure, basins can be defined deterministically
from task-relevant symbolic or structural features. For open-ended reasoning,
basins can instead be defined semantically, using an extracted central
hypothesis and natural language inference (NLI) \citep{balamurali2025revisiting} to group equivalent hypotheses.
This abstraction separates diversity in underlying reasoning strategies from
surface-level variation among trajectories and gives search an explicit notion
of where it has already explored.

\textbf{BASIN.}
We propose \textsc{BASIN}: \textbf{B}asin-\textbf{A}ware \textbf{S}earch for
\textbf{I}nference-time Reaso\textbf{N}ing, a training-free
\emph{history-biased search} principle for inference-time LLM reasoning.
The design is inspired by metadynamics~\citep{laio2002escaping,tiwary2013metadynamics} in molecular
dynamics \citep{wang2020machine}, an enhanced-sampling method that discourages a physical system from
repeatedly visiting previously explored configurations by accumulating a
history-dependent bias. \textsc{BASIN} transfers this principle to reasoning search by tracking how often each reasoning basin has been selected and applying a logarithmic revisit penalty to candidates from over-visited basins, progressively reallocating a fixed inference budget toward underexplored strategies while still allowing strong candidates from familiar basins to remain competitive. Because this history bias can over-explore once search has already found a promising basin, we further introduce \textsc{QA-BASIN}, which weakens the revisit penalty for high-quality basins to better balance exploration and exploitation. Both methods act only at candidate selection, require no training or changes to the underlying generator, and can be incorporated into search procedures that repeatedly select among candidate states.

\textbf{Empirical findings.}
Across various reasoning tasks,
history-biased basin-aware selection improves search under matched inference
budgets. However, greater basin diversity does not always imply higher
accuracy: \textsc{BASIN} helps most when search is trapped in redundant
incorrect regions, but can over-explore after finding a strong basin.
\textsc{QA-BASIN} mitigates this exploration--exploitation trade-off by
preserving high-quality basins, yielding robust gains across models,
tasks, and search frameworks.
We also introduce the \emph{redundancy gap} $\Delta$, which measures the
difference in search concentration between correct and incorrect predictions.
Standard ToT often operates near $\Delta \approx 0$, indicating that correct
and incorrect searches exhibit similar levels of concentration, whereas
\textsc{BASIN} consistently shifts $\Delta$ positive.

Our contributions are threefold. First, we formalize \textbf{reasoning basins}
and identify \textbf{reasoning basin collapse} as a failure mode of
inference-time search. Second, we introduce \textbf{\textsc{BASIN}}, a
training-free history-biased search method, together with its quality-aware
variant \textbf{\textsc{QA-BASIN}}, and show improvements across multiple
domains, models, and search frameworks under fixed inference budgets. Third,
we introduce the \textbf{redundancy gap} $\Delta$ as a diagnostic for
characterizing harmful search redundancy and motivating more adaptive
history-biased reasoning search.
\vspace{-1em}
\section{Related Work}
\label{sec:related_work}

\textbf{Reasoning with intermediate steps.}
Chain-of-thought (CoT) prompting elicits step-by-step reasoning before the final
answer~\citep{wei2022chain}; Zero-shot-CoT shows that a simple ``think step by
step'' prompt can elicit reasoning without demonstrations~\citep{kojima2022large}.
Further methods improve intermediate rationales through structured
decomposition, automatic exemplar selection, and complexity-based
sampling~\citep{zhou2023least,zhang2023automatic,zelikman2022star,fu2023complexity}.
Self-Consistency aggregates multiple sampled traces by majority
vote~\citep{wang2023selfconsistency}. Recent work also explores reasoning in
continuous latent space to improve the flexibility and efficiency of
intermediate computation~\citep{liu2026latent}. These methods improve how
reasoning trajectories are generated, represented, or aggregated, but do not
explicitly track whether search repeatedly revisits the same underlying
reasoning strategy.

\textbf{Search-based reasoning.}
ToT frames LLM reasoning as search over intermediate
states~\citep{yao2023tree}, with extensions to graph-structured reasoning,
planning-style procedures, and agentic tree
search~\citep{besta2024graph,hao2023reasoning,zhou2024language}.
Recent work also improves search efficiency: Dynamic Parallel Tree Search
reduces redundant exploration in ToT-style inference~\citep{ding2025dynamic},
while Policy-Guided Tree Search learns a controller for expansion, branching,
and backtracking~\citep{li2025policy}. Related work adaptively controls
inference compute through early stopping or routing between models with
different reasoning capabilities~\citep{chen2023frugalgpt,zhou2026adaptive,su2026cprouter}.
\textsc{BASIN} is complementary to these approaches: rather than deciding how
long to search, how to expand the tree, or which model to invoke, it introduces
a history-dependent selection bias based on which reasoning strategies have
already been explored. By grouping states into reasoning basins and penalizing
repeated visits, \textsc{BASIN} directly targets reasoning basin collapse
during candidate selection.

\textbf{Reflection, refinement, and verification.}
Self-Refine iteratively improves outputs through feedback and
revision~\citep{madaan2023selfrefine,zhou2026should,zhou2026streamlined}, while Reflexion uses verbal
self-reflection to guide future attempts~\citep{shinn2023reflexion,huang2025beyond}.
Verifier-based approaches rerank candidates using learned or external
evaluators~\citep{lightman2023letsverify}, and recent test-time methods study
adaptive allocation of reasoning effort~\citep{ling2026neural,zhou2026adaptive,
su2026cprouter}. These methods primarily improve trajectory quality, search
control, or inference allocation. \textsc{BASIN} instead controls how search
effort is distributed across underlying reasoning strategies using search
history; \textsc{QA-BASIN} further conditions this history bias on quality so
that promising basins remain competitive while repeatedly visited,
lower-quality basins are discouraged.

\textbf{Diversity-promoting and history-biased search.}
Diverse Beam Search discourages near-duplicate candidates by adding a
diversity penalty across beam groups~\citep{vijayakumar2016diverse}.
At the reasoning level, Diversity of Thought elicits distinct prompt-level
solution approaches~\citep{naik2023diversity}, while Diversity of Thoughts for
agents reduces redundant reflections to broaden decision-space
exploration~\citep{lingam2025enhancing}. \textsc{BASIN} shares the goal of
reducing redundant search but differs in both the unit of diversity and the
mechanism used to enforce it. Rather than maximizing instantaneous diversity
among candidates, it defines equivalence classes of states at the
\emph{strategy level} and accumulates a history-dependent penalty as the same
basin is revisited. This turns diversity from
a local property of the current candidate set into a history-aware search
signal over previously explored reasoning strategies.

\section{Method}
\label{sec:method}

\textsc{BASIN} is a history-biased modification to inference-time reasoning
search. Standard search scores candidate states largely independently of where
search has already spent its compute, even when several candidates pursue the
same underlying reasoning strategy. As a result, substantial inference budget
can be spent repeatedly exploring equivalent trajectories. \textsc{BASIN}
makes this search history explicit by grouping states into \emph{reasoning
basins} and accumulating a revisit penalty over repeatedly selected basins.

\subsection{Reasoning Basins}
\label{sec:basin_definitions}

A \textbf{reasoning basin} is an equivalence class of states that share the
same underlying reasoning strategy. Basin membership captures redundancy
relevant to search rather than surface similarity: states belong to the same
basin when they pursue the same core hypothesis or induce the same relevant
continuation structure, even if their textual realizations differ.

We define a basin assignment function
$\mathcal{B}: \mathcal{S} \rightarrow \mathcal{Z}$ mapping each reasoning
state $s$ to a discrete basin identifier. States satisfying
$\mathcal{B}(s)=\mathcal{B}(s')$ are treated as repeated exploration of the
same strategy. The basin representation is task-dependent: when explicit
structure is available, we use deterministic structural definitions; for
open-ended reasoning, we approximate strategy equivalence semantically.

\textbf{Structural basins.}
For arithmetic tasks such as Game of 24, we define basin membership using the
ordered sequence of operations applied so far and the sorted remaining values:
\begin{equation}
\mathcal{B}(s)=
\bigl(
\mathrm{ops}(s),
\mathrm{sort}(\mathrm{remaining}(s))
\bigr).
\end{equation}
Here $\mathrm{ops}(s)$ is the ordered operator sequence (e.g.,
$(\mathtt{mul},\mathtt{sub})$), so states applying the same operations in a
different order remain distinct. For example, the partial step
\texttt{11,-,1,=,10} on input ${1,11,11,13}$ yields basin
\texttt{sub,|,10,11,13}. This representation captures task-relevant
structural redundancy with lightweight deterministic parsing.

\textbf{Semantic basins.}
For open-ended tasks such as MuSR, exact structural keys are unavailable and
string matching is too brittle. We therefore extract from each reasoning trace
a $\texttt{main\_hypothesis}$, a one-sentence summary of its central claim, and
group states that predict the same answer and express compatible hypotheses
under an NLI model. Concretely, states $s$ and $s'$ are grouped when
\begin{equation}
\mathrm{ans}(s) = \mathrm{ans}(s')
\;\wedge\;
\mathrm{NLI}_{\mathrm{ent}}(h_s,h_{s'}) \geq \tau_e
\;\wedge\;
\mathrm{NLI}_{\mathrm{con}}(h_s,h_{s'}) \leq \tau_c.
\end{equation}
where $h_s$ denotes the extracted hypothesis and
$\tau_e,\tau_c$ control clustering granularity.
We use NLI rather than embedding similarity because reasoning traces often
share substantial narrative context despite supporting different hypotheses;
NLI more directly captures propositional compatibility. We analyze sensitivity
to the extractor, NLI model, and thresholds in
Appendix~\ref{app:semantic_sensitivity}.

\subsection{History-Biased Basin Selection}
\label{sec:dart_penalty}

Let $\mathrm{visits}[z], z\in \mathcal{Z}$ denote the number of times basin
$z=\mathcal{B}(s)$ has previously been selected into the active search set,
and let $f(s)$ be the base score assigned by the underlying search procedure
(e.g., model likelihood, a value estimate, or a heuristic score).
Inspired by metadynamics~\citep{laio2002escaping} in molecular dynamics, which
uses a history-dependent bias to discourage repeated visits to previously
explored regions, \textsc{BASIN} replaces $f(s)$ with
\begin{equation}
\tilde{f}(s)
=
f(s)
-
\lambda
\log\left(
1+\mathrm{visits}[z]
\right).
\label{eq:basin}
\end{equation}
where $\lambda \geq 0$ controls the strength of the history bias.
Visit counts are updated after each selection step. Because all states in the
same basin share a visit count, the penalty accumulates at the strategy level
rather than independently for individual trajectories. Unvisited basins incur
no penalty, while repeated visits receive an increasing but sublinear penalty.
The resulting search is therefore biased by its own history: as a basin is
revisited, candidates from that basin become progressively less competitive,
shifting selection toward underexplored strategies. Importantly, revisits are
not forbidden; a previously explored basin remains selectable whenever its
base-score advantage exceeds the accumulated penalty.

The selection rule is agnostic to how basin membership is constructed:
deterministic structural keys are preferable when available, while semantic
clustering provides a fallback when no exact task-specific equivalence
relation exists.

\subsection{Quality-Aware \textsc{BASIN}}
\label{sec:qa_method}

History-biased exploration introduces an exploration--exploitation trade-off.
Penalizing revisits is useful when search is trapped in a repeatedly explored
weak basin, but can also redirect compute away from a promising basin simply
because it has been visited frequently. When a meaningful quality signal is
available, we therefore introduce \textsc{QA-BASIN}:
\begin{equation}
\tilde{f}(s)
=
f(s)
-
\lambda
\log\left(1+\mathrm{visits}[z]\right)
\left(1-q_{z}\right),
\label{eq:qa_basin}
\end{equation}
where $q_z\in[0,1]$ is the running mean quality score for basin $z$.
The quality term modulates the accumulated history bias: high-quality basins
receive a weaker revisit penalty, whereas low-quality basins approach the
original \textsc{BASIN} penalty. \textsc{QA-BASIN} therefore preserves
promising strategies while continuing to discourage redundant exploration of
weaker ones. Its effectiveness depends on the verifier providing a meaningful
quality signal.

\subsection{Instantiation in ToT}
\label{sec:dart_tot}

We instantiate history-biased basin selection within ToT, although the
principle applies to any inference-time search procedure that maintains
candidate states and repeatedly selects among them. We also evaluate the same
mechanism with Graph of Thoughts (Appendix~\ref{app:got}) and UCT-based MCTS
(Sec.~\ref{sec:generalization}).
Standard ToT maintains a beam $\mathcal{A}_t$ of $k$ active states. At each
round, it expands the active states into candidate continuations, scores them
using $f$, and retains the top $k$. \textsc{BASIN} changes only this selection
step: each candidate is assigned a basin through $\mathcal{B}$, rescored using
Eq.~\eqref{eq:basin} (or Eq.~\eqref{eq:qa_basin} for \textsc{QA-BASIN}), and
ranked by the resulting history-biased score. Visit counts are then updated for
the selected states. All other components of the search remain unchanged.
Thus, \textsc{BASIN} changes not how candidate reasoning states are generated,
but \emph{where search allocates inference compute} as a function of its
exploration history.

\section{Experiments}
\label{sec:experiments}

\subsection{Experimental Setup}
\label{sec:experimental_setup}

\textbf{Datasets.}
Our primary experiments use two complementary benchmarks spanning symbolic and
natural-language reasoning.
\textit{Game of 24} requires combining four integers using basic arithmetic
($+,-,\times,\div$) to obtain 24; we use the standard 100-problem set
from~\citep{yao2023tree}. Solutions are verified exactly by evaluating the
expression and checking number usage.
\textit{MuSR}~\citep{sprague2024musr} is a multi-step reasoning benchmark
covering murder mystery, object placement, and team allocation; we use 300
problems sampled uniformly across subtasks.

To test broader generalization, we additionally evaluate
\textit{HumanEval} \citep{chen2021codex}, \textit{GSM-Hard} \citep{gao2022pal}, and the \emph{Logical Deduction} subtask of BIG-Bench Hard
\citep{srivastava2022bigbench}. HumanEval tests program synthesis
and admits deterministic structural basin definitions, while GSM-Hard tests
challenging mathematical reasoning. Together, these benchmarks span symbolic
search, natural-language reasoning, mathematics, and code generation.

\textbf{Models.}
Our primary Game of 24 experiments use
\texttt{gpt-4o-mini}~\citep{openai2024gpt4o} and
\texttt{Qwen3-27B}~\citep{qwen2025qwen3}; MuSR uses
\texttt{gpt-4o-mini} and \texttt{gpt-oss-120b}~\citep{openai2025gptoss}.
For broader evaluation, we additionally test
\texttt{Qwen2.5-7B-Instruct} \citep{qwen2025qwen25} and \texttt{Llama-3.3-70B-Instruct} \citep{dubey2024llama3}, covering
multiple model families and scales.

\textbf{Search and Basins.}
We use ToT as the primary controlled
search framework. \textsc{BASIN} changes only candidate selection through
Eq.~\eqref{eq:basin}; \textsc{QA-BASIN} uses Eq.~\eqref{eq:qa_basin}.
Generation, search budget, and final-answer selection are otherwise held fixed.
Game of 24 and HumanEval use deterministic structural basin definitions.
For MuSR, we extract a \texttt{main\_hypothesis} from each reasoning trace
and construct semantic basins as described in
\S\ref{sec:basin_definitions}. The same extraction procedure is applied when
computing basin statistics for baseline and basin-aware searches. Under our
nine-round MuSR setup, both conditions use 18 generation and 18 hypothesis
extraction calls per problem. The Appendices~\ref{app:basin_def}-\ref{app:semantic_sensitivity} study
the sensitivity to the extractor and NLI clustering choices.

\textbf{Hyperparameters.}
For Game of 24, we use beam size $k{=}5$, branching factor $b{=}5$, and depth
$T{=}3$. For MuSR, we use beam size $k{=}2$ and nine reasoning rounds.
Semantic clustering uses entailment threshold $\tau_e{=}0.45$ and
contradiction ceiling $\tau_c{=}0.3$, chosen to require moderate positive
support between hypotheses while excluding pairs with substantial
contradictory evidence. Appendix~\ref{app:semantic_sensitivity} shows that
the results are robust to alternative entailment thresholds and semantic
basin constructions. Unless otherwise stated,
$\lambda{=}3.0$. Sampling temperatures are $0.7$ for
Game of 24 and $0.8$ for MuSR and are held fixed across methods.
Appendix~\ref{app:compute} provides compute and implementation details. Appendix \ref{app:prompts} shows prompt templates.

\textbf{Evaluation metrics.}
We report accuracy as the primary performance metric and Pass@k when the final
beam can contain multiple candidate answers. To characterize search structure, we report the number of visited basins $n_{\mathrm{basins}}$ and the effective basin count
$N_{\mathrm{eff}}=\exp(-\sum_z p_z\log p_z)$, where $p_z$ is the fraction
of selected states assigned to basin $z$. $N_{\mathrm{eff}}$ equals the
number of basins under uniform visitation and decreases as search concentrates
on a subset of them, thereby capturing both basin coverage and the evenness of
search allocation. We define redundancy as
$\rho=n_{\mathrm{basins}}-N_{\mathrm{eff}}$ and the redundancy gap as
\begin{equation}
\Delta
=
\mathbb{E}[\rho \mid \mathrm{correct}]
-
\mathbb{E}[\rho \mid \mathrm{incorrect}].
\label{eq:gap_def}
\end{equation}
Thus, $\Delta>0$ indicates that correct searches are more concentrated than
incorrect ones. We use these quantities as diagnostics of search structure
rather than optimization objectives, since greater basin diversity does not
necessarily imply higher accuracy.

\begin{wraptable}{r}{0.58\textwidth}
\vspace{-29pt}
\centering
\caption{Results on Game of 24.
$^{**}p<0.01$, $^{\dagger}p<0.10$.}
\label{tab:game24}
\resizebox{0.58\textwidth}{!}{%
\begin{tabular}{llccc}
\toprule
\textbf{Model} & \textbf{Method}
& \textbf{Acc.}
& $n_{\mathrm{basins}}$
& $\bm{N_{\mathrm{eff}}}$ \\
\midrule
\multirow{2}{*}{\texttt{gpt-4o-mini}}
& ToT
& 0.660
& 27.39
& 26.65 \\
& +\textsc{BASIN}
& \textbf{0.720}$^{\dagger}$
& \textbf{27.94}
& \textbf{27.29} \\
\midrule
\multirow{2}{*}{\texttt{Qwen3-27B}}
& ToT
& 0.430
& 25.64
& 24.88 \\
& +\textsc{BASIN}
& \textbf{0.650}$^{**}$
& \textbf{28.15}
& \textbf{27.38} \\
\bottomrule
\end{tabular}}
\vspace{-8pt}
\end{wraptable}
\subsection{Main Results}
\label{sec:main_results}

\textbf{Game of 24.}
Table~\ref{tab:game24} shows that \textsc{BASIN} improves accuracy from
66.0\% to 72.0\% with \texttt{gpt-4o-mini}, and from 43.0\% to
65.0\% with \texttt{Qwen3-27B} ($+22$pp, $p{<}0.01$).
The gains occur under the same search budget and exact symbolic verifier.
\textsc{BASIN} also increases $N_{\mathrm{eff}}$, particularly for
\texttt{Qwen3-27B}, suggesting that ToT spends substantial compute
revisiting structurally redundant arithmetic states. We observe similar findings for the BBH logical deduction task (Appendix \ref{app:bbh}) with +13pp in accuracy.

\textbf{MuSR.}
Table~\ref{tab:musr} reports results on 300 MuSR problems. The NLI-based semantic
construction is inherently noisier than the exact structural basin definition
used for Game of 24, which may limit how precisely the revisit penalty distinguishes genuinely different reasoning strategies. 
Unlike Game of 24, global diversity changes on MuSR are small. For
\texttt{gpt-oss-120b}, \textsc{BASIN} improves both accuracy and Pass@k;
for \texttt{gpt-4o-mini}, Pass@k decreases slightly while accuracy increases.
We define selection efficiency as $\mathrm{Acc.}/\mathrm{Pass@}k$;
\textsc{BASIN} achieves the highest selection efficiency for both models.
This suggests that BASIN performance depends not only on exploration but also on the
quality of the semantic basin representation. Appendix~\ref{app:semantic_sensitivity}
shows that accuracy remains stable across alternative semantic constructions.
Overall, the gains arise from reallocating search across strategies rather than
simply maximizing basin coverage.

\textbf{Quality-aware selection.}
We further investigate the exploration--exploitation trade-off in BASIN. Table~\ref{tab:qa_basin} compares standard ToT,
\textsc{BASIN}, and \textsc{QA-BASIN} using
\texttt{gpt-4}~\citep{achiam2023gpt}. On Game of 24, standard ToT obtains 67.0\% accuracy, flat
\textsc{BASIN} falls to 61.0\%, and \textsc{QA-BASIN} reaches 70.0\%.
On MuSR, the corresponding accuracies are 52.0\%, 58.3\%, and 58.7\%.
Notably, on Game of 24 the three methods achieve nearly identical effective
basin counts despite substantially different accuracies. This reinforces that
the objective is not to maximize basin diversity itself, but to avoid
redundant exploration without suppressing promising reasoning regions.
\textsc{QA-BASIN} directly addresses this trade-off by weakening the revisit
penalty for basins with stronger quality signals.

\begin{wraptable}{r}{0.58\textwidth}
\vspace{-20pt}
\centering
\caption{\textsc{QA-BASIN} vs.\ baselines on \texttt{gpt-4}.}
\label{tab:qa_basin}
\small
\begin{tabular}{llccc}
\toprule
\textbf{Task} & \textbf{Method}
& \textbf{Acc.} & $n_{\mathrm{basins}}$ & $\bm{N_{\mathrm{eff}}}$ \\
\midrule
\multirow{3}{*}{Game24}
& ToT
& 0.670
& 28.17
& 27.65 \\
& \textsc{BASIN}
& 0.610
& 28.24
& 27.72 \\
& \textsc{QA-BASIN}
& \textbf{0.700}
& \textbf{28.25}
& \textbf{27.73} \\
\midrule
\multirow{3}{*}{MuSR}
& ToT
& 0.520
& 5.86
& 4.67 \\
& \textsc{BASIN}
& 0.583
& \textbf{7.39}
& \textbf{5.72} \\
& \textsc{QA-BASIN}
& \textbf{0.587}
& 7.32
& 5.64 \\
\bottomrule
\end{tabular}
\vspace{-10pt}
\end{wraptable}

\subsection{Generalization Across Tasks, Models, and Search}
\label{sec:generalization}

We next test whether basin-aware selection generalizes beyond the primary
Game of 24 and MuSR settings. Table~\ref{tab:generalization} reports accuracy
under matched inference budgets on HumanEval and GSM-Hard across four models.
Across these settings, \textsc{QA-BASIN} is generally the most robust
formulation: it matches or improves upon standard ToT for all four models on
HumanEval and achieves the best or tied-best accuracy in three of four
GSM-Hard settings. In contrast, flat \textsc{BASIN} sometimes increases
exploration without improving accuracy, consistent with an
exploration--exploitation trade-off rather than a simple benefit from greater
basin coverage. This pattern also extends to additional Game of 24 models:
both basin-aware variants improve over ToT on Qwen2.5-7B-Instruct, while
\textsc{QA-BASIN} improves Llama-3.3-70B-Instruct from $70\%$ to $75\%$
(Appendix~\ref{app:game24_additional_models}).

\begin{table}[t]
\centering
\small
\caption{Results on MuSR ($n{=}300$, nine rounds, $\lambda{=}3.0$). Sel. Eff. indicates selection efficiency.}
\label{tab:musr}
\begin{tabular}{llccccc}
\toprule
\textbf{Model} & \textbf{Method}
& \textbf{Acc.} & \textbf{Pass@k} & \textbf{Sel. Eff.}
& $n_{\mathrm{basins}}$ & $\bm{N_{\mathrm{eff}}}$ \\
\midrule
\multirow{2}{*}{\texttt{gpt-oss-120b}}
& ToT
& 0.633 & 0.863 & 0.734 & \textbf{8.59} & \textbf{6.86} \\
& +\textsc{BASIN}
& \textbf{0.670}$^{*}$ & \textbf{0.903} & \textbf{0.742}
& 8.58 & 6.85 \\
\midrule
\multirow{2}{*}{\texttt{gpt-4o-mini}}
& ToT
& 0.607 & \textbf{0.857} & 0.708 & 6.76 & 5.07 \\
& +\textsc{BASIN}
& \textbf{0.620}$^{\dagger}$ & 0.833 & \textbf{0.744}
& \textbf{6.84} & \textbf{5.28} \\
\bottomrule
\end{tabular}
\end{table}

\begin{table}[t]
\centering
\small
\caption{\textbf{Generalization across tasks and models.}
Accuracy under matched inference budgets.}
\label{tab:generalization}
\begin{tabular}{llccc}
\toprule
\textbf{Task} & \textbf{Model}
& \textbf{ToT}
& \textbf{\textsc{BASIN}}
& \textbf{\textsc{QA-BASIN}} \\
\midrule
HumanEval
& gpt-4o-mini
& .799
& .811
& \textbf{.817} \\
& gpt-oss-120b
& .756
& .750
& \textbf{.793} \\
& Qwen2.5-7B-Instruct
& .756
& .750
& \textbf{.780} \\
& Llama-3.3-70B-Instruct
& \textbf{.817}
& .799
& \textbf{.817} \\
\midrule
GSM-Hard
& gpt-4o-mini
& .510
& \textbf{.530}
& \textbf{.530} \\
& gpt-oss-120b
& \textbf{.610}
& \textbf{.610}
& \textbf{.610} \\
& Qwen2.5-7B-Instruct
& .440
& .450
& \textbf{.460} \\
& Llama-3.3-70B-Instruct
& .440
& \textbf{.470}
& .440 \\
\bottomrule
\end{tabular}
\end{table}

\begin{wraptable}{r}{0.58\textwidth}
\vspace{-29pt}
\centering
\caption{Game of 24 with UCT-based MCTS.}
\label{tab:mcts_main}
\small
\begin{tabular}{lccc}
\toprule
\textbf{Model}
& \textbf{MCTS}
& \textbf{\textsc{BASIN}}
& \textbf{\textsc{QA-BASIN}} \\
\midrule
gpt-4o-mini
& .460
& .420
& \textbf{.640} \\
gpt-oss-120b
& .240
& .270
& \textbf{.390} \\
Qwen2.5-7B-Instruct
& .490
& .360
& \textbf{.550} \\
Llama-3.3-70B-Instruct
& .610
& .650
& \textbf{.720} \\
\bottomrule
\end{tabular}
\vspace{-10pt}
\end{wraptable}

\textbf{Generalization beyond ToT.}
Because \textsc{BASIN} modifies candidate selection rather than the topology
of a particular search algorithm, we also evaluate it with UCT-based Monte
Carlo Tree Search (MCTS) on Game of 24 under matched search budgets. We use 50 simulations per problem and
add the basin term only to UCT child selection (Table \ref{tab:mcts_main}).

\textsc{QA-BASIN} improves over standard MCTS for all four models, by
$+18$, $+15$, $+6$, and $+11$ pp, respectively. Flat
\textsc{BASIN}, however, helps two models and hurts two. Since MCTS already
contains an explicit exploration term, an additional unconditional revisit
penalty can over-explore and displace promising regions. The quality-aware
variant instead preserves high-quality basins while discouraging repeated
visits to weaker ones. We observe the same transfer beyond tree search with Graph of Thoughts (GoT)
on MuSR: \textsc{BASIN} improves accuracy from $57\%$ to $60\%$, while
\textsc{QA-BASIN} further improves it to $64\%$. Notably,
\textsc{QA-BASIN} achieves this gain with lower effective basin coverage than
flat \textsc{BASIN}, again illustrating that effective basin-aware search
requires balancing exploration with preservation of promising reasoning
regions. Full results are reported in Appendix~\ref{app:got}.

Taken together, these results show that basin-aware selection transfers
across tasks, model families, and search algorithms. They also motivate
\textsc{QA-BASIN} as the preferred formulation when a reliable quality signal
is available: it retains the benefit of escaping repeatedly explored
low-quality basins while reducing the over-exploration that can arise from
flat \textsc{BASIN}.
\subsection{Collapse-Stratified Analysis}
\label{sec:collapse_stratified}

\begin{figure}[t]
\centering
\includegraphics[width=\linewidth]{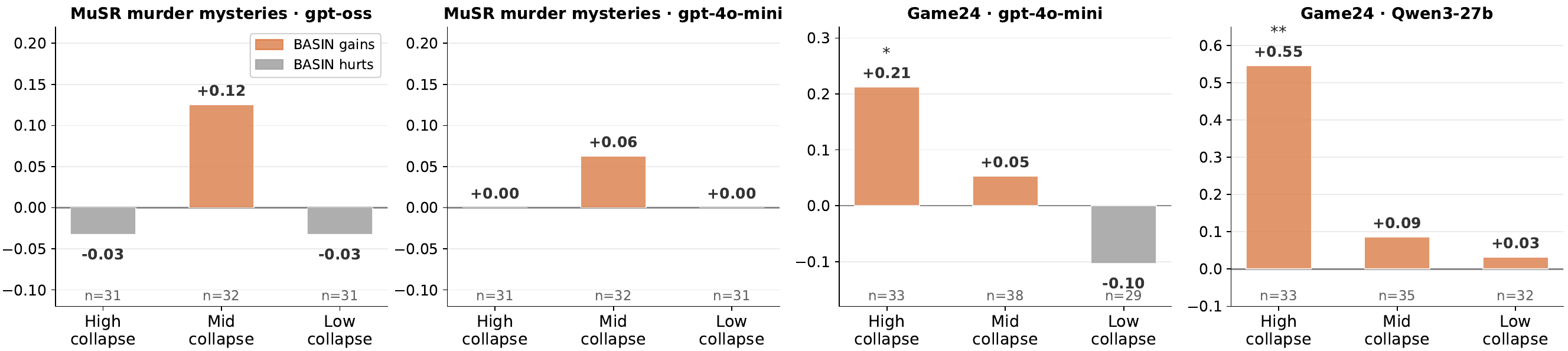}
\caption{
\textsc{BASIN} accuracy gain stratified by ToT basin collapse.
Problems are grouped into tertiles by ToT $N_{\mathrm{eff}}$.
$^{**}p<0.01$, $^{*}p<0.05$ (one-sided McNemar test).
}
\label{fig:collapse_supplement}
\vspace{-5pt}
\end{figure}

If reasoning basin collapse is an important failure mode, \textsc{BASIN} should help
most when standard ToT repeatedly concentrates on a small set of strategies.
We test this by splitting problems into tertiles according to standard-ToT
$N_{\mathrm{eff}}$: high-, mid-, and low-collapse. We then report paired
accuracy differences
$\Delta\mathrm{acc}
=\mathrm{acc}_{\textsc{BASIN}}-\mathrm{acc}_{\mathrm{ToT}}$ in Figure \ref{fig:collapse_supplement}.
For MuSR, we use the murder-mystery subset to avoid mixing heterogeneous
subtasks; Appendix~\ref{app:musr_subtasks} reports the remaining subsets.

The expected pattern is clearest on Game of 24. With
\texttt{gpt-4o-mini}, \textsc{BASIN} gains $+0.21$ accuracy in the
high-collapse group, compared with $+0.05$ in the mid-collapse group and
$-0.10$ in the low-collapse group. \texttt{Qwen3-27B} shows the same
qualitative trend, with most of the improvement concentrated among
high-collapse problems.

MuSR is less monotonic: the largest gains occur in the mid-collapse group.
Semantic $N_{\mathrm{eff}}$ captures the amount of concentration but not
whether the dominant reasoning basin is useful or misleading. This weaker
alignment between collapse severity and the need for exploration may partly
explain the smaller gains on MuSR relative to Game of 24. Overall, collapse
severity is informative but insufficient by itself to determine when
additional exploration will help.

\subsection{Understanding Search Through the Redundancy Gap}
\label{sec:gap_metric}

The flat \textsc{BASIN} penalty is quality-agnostic: it depends on basin
visitation rather than correctness. We use the redundancy gap $\Delta$ from
Eq.~\eqref{eq:gap_def} to analyze how the resulting concentration differs
between successful and unsuccessful searches.

\begin{wrapfigure}{r}{0.5\linewidth}
\centering
\includegraphics[width=\linewidth]{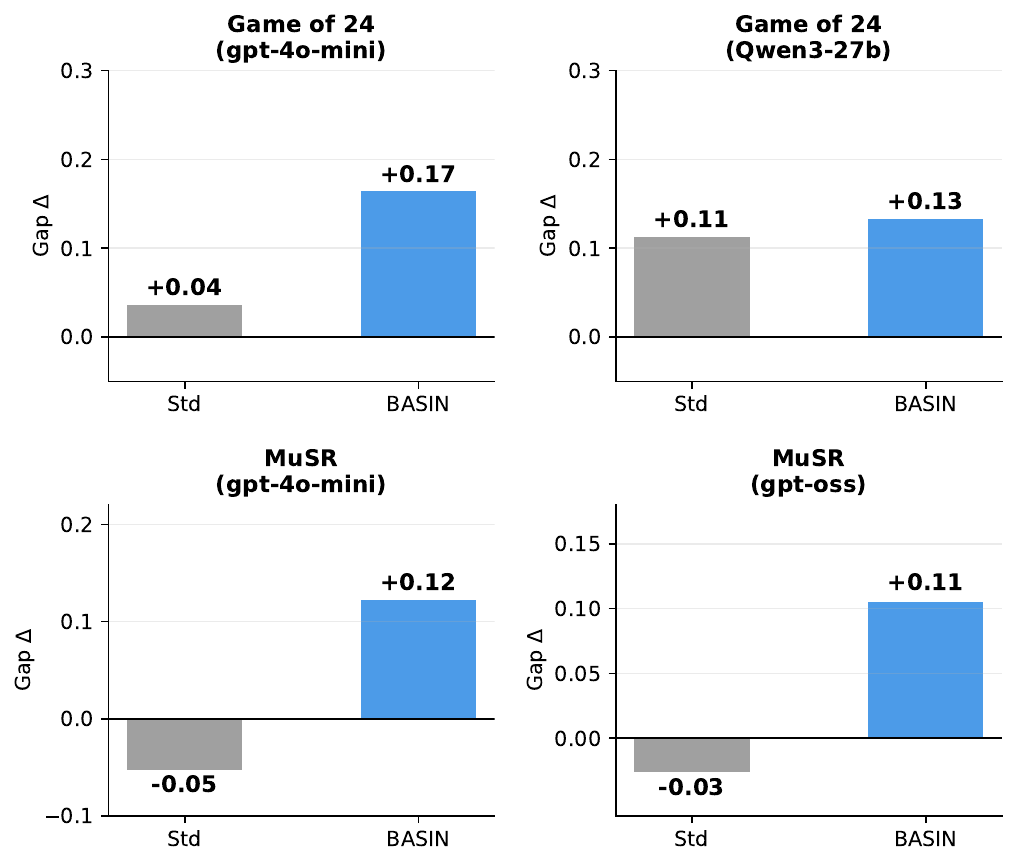}
\vspace{-10pt}
\caption{Redundancy gap $\Delta$ for ToT and
\textsc{BASIN}.}
\label{fig:gap_bar}
\vspace{-10pt}
\end{wrapfigure}

Figure~\ref{fig:gap_bar} shows that standard ToT (Std) typically operates near
$\Delta\approx0$ or slightly below: correct and incorrect searches exhibit
similar concentration patterns. \textsc{BASIN} consistently shifts $\Delta$
positive, concentrating successful searches around strong regions while
dispersing repeated exploration among unsuccessful ones. This separation helps
explain why a quality-agnostic revisit penalty can improve accuracy: its effect
depends not simply on increasing diversity, but on restructuring where
redundancy occurs.

The relationship is useful but not universal. For
\texttt{Qwen3-27b} on Game of 24, ToT already has $\Delta=+0.11$, yet
\textsc{BASIN} improves accuracy by +22pp. Since ToT solves only 43\% of
problems in this setting, substantial per-problem collapse onto incorrect
arithmetic basins can remain even when the dataset-level gap appears favorable.
Thus, $\Delta$ summarizes average search behavior but can obscure substantial
per-problem heterogeneity.

The redundancy gap therefore provides a useful summary of how search
concentration differs between successful and unsuccessful trajectories, but it
does not by itself determine when additional exploration will improve
accuracy. We therefore evaluate whether $\Delta$
can be used as a routing signal for choosing between standard search and
\textsc{BASIN}. Across six model--task
settings spanning Game of 24 and BBH, $\Delta$ alone selects the empirically
better fixed policy in only $2/6$ cases. Combining it
with a per-problem search-effort signal---the number of tree nodes explored
before termination---increases this to $5/6$; Appendix~\ref{app:gap_routing}
provides the full routing analysis. Thus, we treat the redundancy gap primarily as a
diagnostic of search structure, while adaptive policies should combine it with
additional search-state or quality signals.

\subsection{Ablation Studies}
\label{sec:ablations}

\begin{wrapfigure}{r}{0.4\textwidth}
\vspace{-12pt}
\centering
\includegraphics[width=0.4\textwidth]{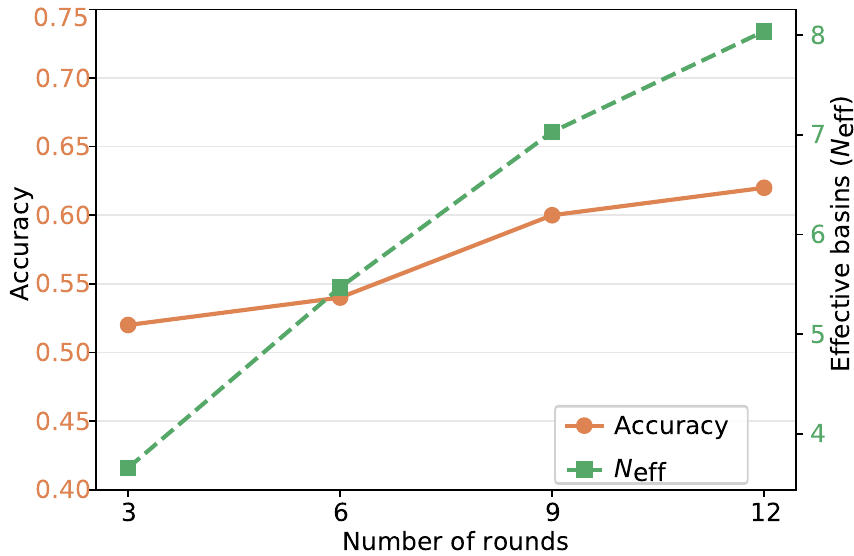}
\caption{Effect of compute budget.}
\label{fig:rounds_ablation}
\vspace{-15pt}
\end{wrapfigure}

\textbf{Effect of compute budget.}
Figure~\ref{fig:rounds_ablation} varies the number of reasoning rounds on 100
randomly selected MuSR problems with \texttt{gpt-oss-120b}, holding beam size
fixed. Accuracy generally improves with additional reasoning depth, while
$N_{\mathrm{eff}}$ also increases. Basin-aware exploration is therefore most
useful when the search budget is large for alternative strategies to
develop into complete solutions.

\begin{figure}
\vspace{-15pt}
\centering
\includegraphics[width=0.8\textwidth]{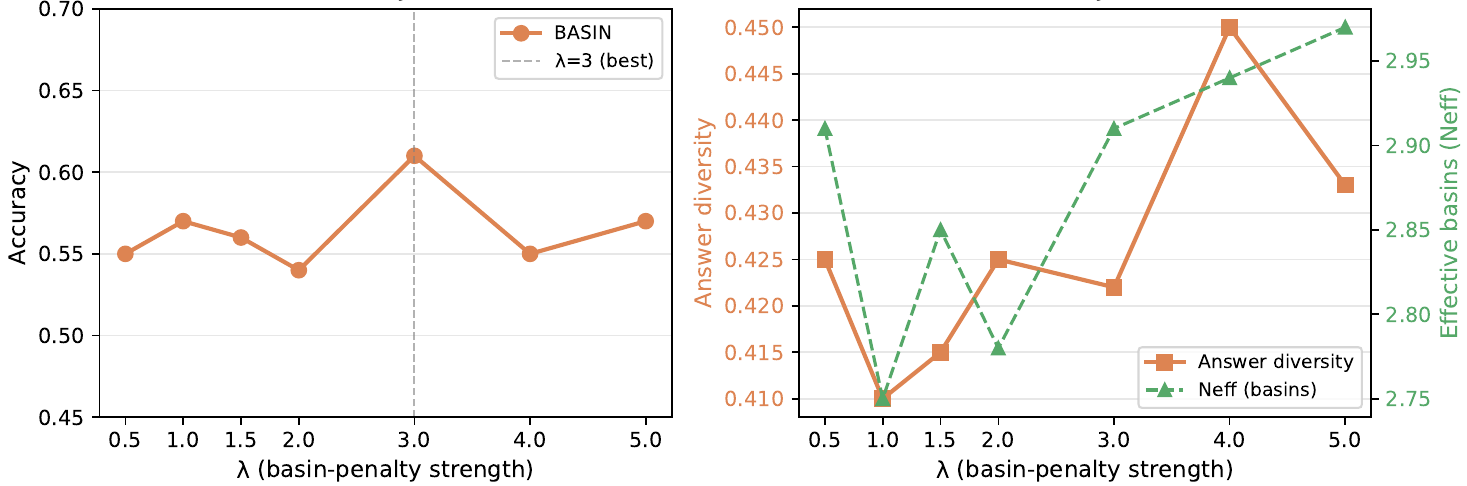}
\vspace{-10pt}
\caption{Effect of penalty strength.}
\label{fig:lambda_sweep}
\vspace{-15pt}
\end{figure}

\textbf{Effect of penalty strength.}
Figure~\ref{fig:lambda_sweep} varies
$\lambda\in\{0.5,1.0,1.5,2.0,3.0,4.0,5.0\}$ in the same MuSR setting.
Accuracy peaks at $\lambda{=}3.0$ and is relatively stable over moderate
values. $N_{\mathrm{eff}}$ increases with $\lambda$, but accuracy is
non-monotonic: weak penalties have little effect, whereas overly strong
penalties can override useful base-score differences. This again shows that
the goal is not maximal diversity, but an effective exploration--exploitation
balance.

\textbf{How important is the quality signal?}
Because \textsc{QA-BASIN} uses basin quality to modulate the revisit penalty,
its performance depends on the informativeness of that signal. On MuSR with
\texttt{gpt-4}, using an LLM-based quality signal yields 58.7\% accuracy,
compared with 58.3\% for flat \textsc{BASIN} and 52.0\% for standard ToT.
Replacing this signal with the search heuristic reduces accuracy to 33.7\%.
Thus, quality-aware modulation is beneficial when the quality estimate is
informative, but can be actively harmful when it is poorly calibrated to
correctness. We therefore recommend \textsc{QA-BASIN} when a meaningful
quality signal is available and flat \textsc{BASIN} otherwise.
Appendix~\ref{app:qa_basin_verifier} provides the full results and analyzes
the discriminative quality of the heuristic score.

\begin{wrapfigure}{r}{0.4\textwidth}
  \vspace{-12pt}
  \centering
  \includegraphics[width=0.4\textwidth]{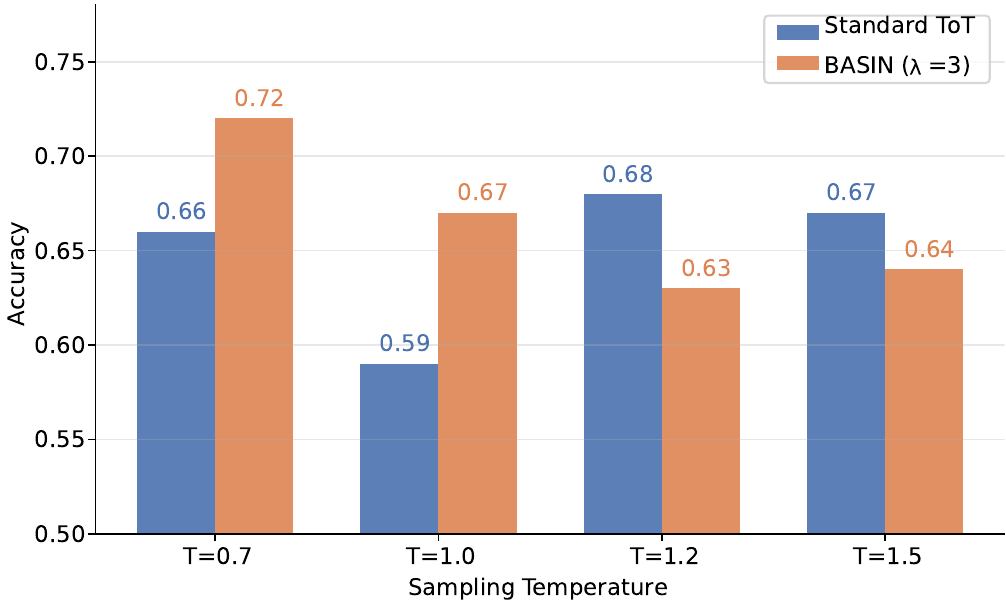}
    \vspace{-10pt}
  \caption{Effect of temperature $T$.}
  \label{fig:temperature_ablation}
  \vspace{-10pt}
\end{wrapfigure}

\textbf{Is \textsc{BASIN} Just Promoting Diversity?}
The improvement from \textsc{BASIN} is not explained by generic diversity
promotion. On Game of 24 with \texttt{gpt-4o-mini}, a Diverse Beam Search
(DBS)-style baseline~\citep{vijayakumar2016diverse} achieves 64.0\% accuracy,
compared with 66.0\% for standard ToT and 72.0\% for \textsc{BASIN}. Thus,
encouraging diverse candidates alone does not reproduce the gain from explicitly
modeling strategy-level redundancy.
We further vary the sampling temperature as an alternative way to increase
token-level diversity (Fig. \ref{fig:temperature_ablation}). Across $T\in\{0.7,1.0,1.2,1.5\}$, the best
higher-temperature ToT configuration reaches 68.0\% accuracy, still below
\textsc{BASIN}'s 72.0\% at $T=0.7$. Increasing temperature within
\textsc{BASIN} also does not improve performance. These results show
that token-level or candidate-level diversity is not interchangeable with
history-biased, basin-level selection: \textsc{BASIN} benefits from identifying
and penalizing repeated reasoning strategies rather than simply making
individual candidates more different.

\subsection{Case Study}
\label{sec:case_studies}

We illustrate how \textsc{BASIN} reallocates search effort on Game of 24; a
MuSR example appears in Appendix~\ref{app:example}.

\textbf{Game of 24: ${1,11,11,13}$.}
Under standard ToT, the first-step beam contains the same symbolic state twice:
\begin{center}
\ttfamily
\small
11 - 1 = 10,\quad
13 - 11 = 2,\quad
11 * 1 = 11,\quad
11 + 11 = 22,\quad
\textbf{11 - 1 = 10}.
\end{center}
The first and final candidates both map to basin
\texttt{sub|10,11,13}, so one beam slot is spent revisiting the same
arithmetic state. After its first visit, \textsc{BASIN} lowers the score of
this basin, allowing the alternative $1+11=12$ to survive. This new basin
leads to

$$
1+11=12
\;\longrightarrow\;
13-11=2
\;\longrightarrow\;
2\times12=24.
$$

No new generator, operation, or verifier is introduced; the search simply
allocates its existing beam budget across distinct symbolic strategies.

\section{Discussion}

\textbf{Why history-biased basin search can improve reasoning.}
\textsc{BASIN} is motivated by the observation that inference-time reasoning
search can over-commit to a small number of plausible but incomplete reasoning
directions. In ToT-style search, early selections shape later expansions: if
the beam repeatedly selects variants of the same hypothesis, subsequent rounds
tend to elaborate that hypothesis rather than test alternatives. By tracking
which reasoning basins have already been explored and penalizing repeated
visits, \textsc{BASIN} introduces a history-dependent bias that reallocates
search effort toward underexplored strategies and reduces the risk that all
active states inherit the same error mode.

This mechanism is not equivalent to maximizing diversity. The Diverse Beam
Search and temperature comparisons in \S\ref{sec:ablations} show that generic
candidate- or token-level diversity does not reproduce the gains from modeling
strategy-level redundancy, and increasing $N_{\mathrm{eff}}$ can coincide with
unchanged or lower accuracy. The objective is therefore not maximal diversity,
but avoiding excessive reuse of the same underlying reasoning strategy.

The redundancy gap $\Delta$ (\S\ref{sec:gap_metric}) provides a diagnostic of
this behavior. Standard ToT typically operates near $\Delta \approx 0$,
whereas \textsc{BASIN} often shifts the gap positive. However, $\Delta$ alone
selects the empirically better fixed policy in only $2/6$ settings; combining
it with a search-effort signal succeeds in $5/6$ settings
(Appendix~\ref{app:gap_routing}). We therefore treat $\Delta$ as a diagnostic
of harmful concentration rather than a standalone routing criterion.

\textbf{Exploration versus exploitation.}
Flat \textsc{BASIN} is most useful when baseline search repeatedly commits
compute to an unproductive basin, but it can hurt when search has already
identified a promising one. This effect is especially visible with stronger
models and under MCTS, where an existing exploration mechanism can compound the
additional history-based exploration pressure. The collapse-stratified results
similarly show that additional exploration is most useful when baseline search
is sufficiently concentrated.
\textsc{QA-BASIN} addresses this trade-off by weakening the revisit penalty for
basins with stronger quality evidence. Across tasks and models, it is more
robust than the unconditional penalty, and under MCTS it improves over standard
search for all four evaluated models. We therefore view \textsc{QA-BASIN} as
the preferred formulation when a meaningful quality signal is available, with
flat \textsc{BASIN} providing a simpler alternative when it is not.

\textbf{The importance of basin structure.}
The effectiveness of history-biased search depends on the basin representation
capturing meaningful reasoning redundancy. For symbolic tasks, basin membership
can often be defined exactly; for open-ended reasoning, it must be
approximated. On MuSR, SBERT-based similarity collapses trajectories into
nearly a single basin ($N_{\mathrm{eff}} \approx 1.04$), whereas NLI-based
clustering over extracted hypotheses yields a richer strategy-level structure
($N_{\mathrm{eff}} \approx 6.85$; Appendix~\ref{app:basin_def}). At the same
time, performance is robust within this semantic construction family: accuracy
remains above standard ToT across tested entailment thresholds, NLI models, and
hypothesis extractors, although basin counts and Pass@k vary more
(Appendix~\ref{app:semantic_sensitivity}).

\textbf{Generality and limitations.}
Results across HumanEval, GSM-Hard, Game of 24, MuSR, MCTS, and GoT suggest
that reasoning-basin structure is not tied to a single task, model family, or
search topology. More broadly, \textsc{BASIN} suggests \emph{history-biased
reasoning search} as a general principle: represent strategy-level redundancy
through reasoning basins, then use accumulated search history to influence
where inference compute is allocated.

Several limitations remain. Basin definitions are task-dependent, and semantic
tasks require approximate representations with additional extraction and NLI
cost. Neither basin coverage nor $N_{\mathrm{eff}}$ determines whether the
explored strategy is correct, while \textsc{QA-BASIN} additionally depends on
the reliability of its quality signal. Our experiments also use matched
inference budgets rather than characterizing the full accuracy--token-cost
Pareto frontier. Finally, the redundancy gap is insufficient by itself for
deciding when exploration is beneficial; adaptive policies combining basin
visitation, quality estimates, and online search-state signals are a promising
direction for future work.
\section*{Acknowledgments}
This work is supported by the National Science Foundation (NSF) Grant \#2312862, NSF-Simons SkAI Institute, NSF CAREER \#2440542, NSF \#2533996, NSF \#2621883, National Institutes of Health (NIH) \#R01AG091762, NSF ACCESS Computing Resources, NAIRR, NRP, a Google Research Scholar Award, and Cisco gift grant.
\bibliographystyle{unsrtnat}
\bibliography{ref}
\appendix

\section{MuSR Per-Subtask Results}
\label{app:musr_subtasks}

MuSR~\citep{sprague2024musr} contains three reasoning subtasks: murder
mysteries, team allocation, and object placement. Table~\ref{tab:musr_subtasks}
reports results across all three subtasks and both models. These results provide
a finer-grained view of when a quality-agnostic basin-revisit penalty helps and
when additional exploration can instead disrupt a promising reasoning region.

\begin{table}[h]
\centering
\small
\caption{\textsc{BASIN} vs.\ ToT on all three MuSR subtasks.
$\Delta_{Acc}$ denotes the \textsc{BASIN} minus ToT accuracy difference.
$N_{\mathrm{eff}}$ is the mean effective basin count under standard ToT and $n$ is the sample size.
Pass@$k$ is the fraction of problems for which at least one final candidate
is correct.}
\label{tab:musr_subtasks}
\begin{tabular}{llrccccr}
\toprule
\textbf{Model} & \textbf{Task} & $\bm{n}$ & \textbf{ToT}
& \textbf{\textsc{BASIN}} & $\bm{\Delta}_{Acc}$
& \textbf{Pass@$k$ (ToT / \textsc{BASIN})}
& $\bm{N_{\mathrm{eff}}}$ \\
\midrule
\multirow{3}{*}{\texttt{gpt-oss-120b}}
& Murder & 94
& 0.628 & \textbf{0.649} & $+0.021$
& 0.915 / \textbf{0.957} & 8.06 \\
& Team & 96
& \textbf{0.719} & \textbf{0.719} & $+0.000$
& 0.938 / \textbf{0.958} & 8.76 \\
& Object & 110
& 0.564 & \textbf{0.645} & $+0.082$
& 0.755 / \textbf{0.809} & 4.18 \\
\midrule
\multirow{3}{*}{\texttt{gpt-4o-mini}}
& Murder & 94
& 0.649 & \textbf{0.670} & $+0.021$
& \textbf{0.883} / 0.862 & 6.39 \\
& Team & 96
& 0.562 & \textbf{0.625} & $+0.062$
& \textbf{0.865} / 0.812 & 5.18 \\
& Object & 110
& \textbf{0.609} & 0.573 & $-0.036$
& \textbf{0.827} / \textbf{0.827} & 3.84 \\
\bottomrule
\end{tabular}
\end{table}

The results reinforce that basin concentration alone does not determine whether
additional exploration will help. Low $N_{\mathrm{eff}}$ can indicate harmful
collapse onto an incorrect strategy, but it can also reflect useful agreement
around a strong solution. Pass@$k$, final accuracy, and the nature of the
dominant basin are therefore important for interpreting the effect of
\textsc{BASIN}.

\textbf{Murder mysteries.}
Both models show a small positive accuracy gain
(\texttt{gpt-oss-120b}: $+0.021$, 9 wins vs.\ 7 losses;
\texttt{gpt-4o-mini}: $+0.021$, 7 wins vs.\ 5 losses), although neither
difference is statistically significant ($p=0.40$ and $p=0.39$,
respectively). The collapse-stratified analysis in
\S\ref{sec:collapse_stratified} further shows that the effect is heterogeneous
across problems rather than uniformly determined by aggregate basin diversity.
For \texttt{gpt-oss-120b}, standard ToT already achieves Pass@$k=0.915$,
leaving limited room to improve candidate coverage. For
\texttt{gpt-4o-mini}, Pass@$k=0.883$ similarly indicates that a substantial
part of the remaining error comes from selecting among already discovered
candidates rather than from failing to explore the correct answer entirely.

\textbf{Object placement.}
The two models behave differently despite both exhibiting relatively low
effective basin counts. For \texttt{gpt-oss-120b}, standard ToT has
$N_{\mathrm{eff}}=4.18$ and Pass@$k=0.755$, so nearly one quarter of
problems contain no correct final candidate. \textsc{BASIN} improves accuracy
by $+0.082$ ($p=0.032$, 14 wins vs.\ 5 losses), consistent with additional
strategy-level exploration recovering answers that standard search misses.

For \texttt{gpt-4o-mini}, standard ToT has an even lower
$N_{\mathrm{eff}}=3.84$, yet Pass@$k$ is already $0.827$ and
\textsc{BASIN} reduces final accuracy by $0.036$. Thus, low effective basin
count is not by itself evidence of harmful collapse. In this setting,
additional exploration can displace a useful consensus rather than recover a
missing reasoning strategy.

\textbf{Team allocation.}
\texttt{gpt-4o-mini} gains $+0.062$ accuracy ($p=0.035$), whereas
\texttt{gpt-oss-120b} is unchanged. The latter already exhibits high
effective basin coverage ($N_{\mathrm{eff}}=8.76$), while
\texttt{gpt-4o-mini} operates at $N_{\mathrm{eff}}=5.18$.
The result is consistent with \textsc{BASIN} being most useful when the
baseline search underexplores materially different strategies, but the object
placement results above show why diversity statistics alone are insufficient
to identify that regime.

\textbf{Summary.}
Across the six task--model combinations, the two statistically significant
positive results occur in settings where additional strategy-level exploration
can recover or preserve useful alternatives. The negative result demonstrates
the complementary failure mode: an unconditional revisit penalty can
over-explore when concentration reflects useful exploitation rather than
pathological collapse. This motivates the quality-aware formulation in
Eq.~\eqref{eq:qa_basin}, which weakens the penalty for high-quality basins.

\section{Additional Game of 24 Models}
\label{app:game24_additional_models}

To complement the main Game of 24 results, we evaluate
Qwen2.5-7B-Instruct and Llama-3.3-70B-Instruct under the same matched-budget
protocol. Table~\ref{tab:game24_additional_models} reports accuracy for
standard ToT, \textsc{BASIN}, and \textsc{QA-BASIN}.

\begin{table}[h]
\centering
\small
\caption{\textbf{Additional Game of 24 model results.}
Accuracy under matched inference budgets. Bold denotes the best result within
each model.}
\label{tab:game24_additional_models}
\begin{tabular}{lccc}
\toprule
\textbf{Model}
& \textbf{ToT}
& \textbf{\textsc{BASIN}}
& \textbf{\textsc{QA-BASIN}} \\
\midrule
Qwen2.5-7B-Instruct
& 0.600
& \textbf{0.650}
& 0.640 \\
Llama-3.3-70B-Instruct
& 0.700
& 0.680
& \textbf{0.750} \\
\bottomrule
\end{tabular}
\end{table}

The additional models exhibit the same exploration--exploitation trade-off
observed elsewhere. Flat \textsc{BASIN} improves Qwen2.5-7B-Instruct from
$60\%$ to $65\%$, but slightly reduces accuracy for Llama-3.3-70B-Instruct
from $70\%$ to $68\%$. In contrast, \textsc{QA-BASIN} reaches $75\%$ on
Llama-3.3-70B-Instruct, illustrating the benefit of preserving promising
basins when unconditional exploration can displace high-quality trajectories.
\section{BBH Logical Deduction}
\label{app:bbh}

To evaluate transfer beyond the main benchmarks, we apply \textsc{BASIN} to
the \emph{Logical Deduction} subtask of BIG-Bench Hard
\citep{srivastava2022bigbench}. The task requires ordering objects from
relational constraints and therefore differs from both arithmetic search and
open-ended abductive reasoning. We evaluate $n=100$ problems with
\texttt{gpt-4o-mini} and $\lambda=3.0$.

\begin{table}[h]
\centering
\small
\caption{Results on BBH Logical Deduction
(\texttt{gpt-4o-mini}, $n{=}100$, $\lambda{=}3.0$).
$^{*}p<0.05$ using a one-sided McNemar test.}
\label{tab:bbh}
\begin{tabular}{lccc}
\toprule
\textbf{Method}
& \textbf{Acc.}
& $n_{\mathrm{basins}}$
& $\bm{N_{\mathrm{eff}}}$ \\
\midrule
Standard ToT
& 0.400
& 6.49
& 4.50 \\
ToT + \textsc{BASIN}
& \textbf{0.530}$^{*}$
& \textbf{7.11}
& \textbf{5.60} \\
\bottomrule
\end{tabular}
\end{table}

\textsc{BASIN} improves accuracy by $13$pp over standard ToT
($p=0.018$, 23 wins vs.\ 10 losses) while increasing
$N_{\mathrm{eff}}$ from $4.50$ to $5.60$. This result provides an additional
example in which reallocating search toward distinct reasoning strategies
improves accuracy. As elsewhere in the paper, however, the increase in
$N_{\mathrm{eff}}$ should be interpreted as a description of the changed
search behavior rather than as the objective itself.

\section{Graph-of-Thought Backbone}
\label{app:got}

We additionally evaluate \textsc{BASIN} and \textsc{QA-BASIN} with a
Graph-of-Thought (GoT) backbone~\citep{besta2024graph} on MuSR
(\texttt{gpt-oss-120b}, $n{=}100$, 12 calls per problem). GoT extends ToT
with an explicit aggregation step that merges top-scoring trajectories into a
refined answer. We apply the basin-revisit penalty during search selection
while leaving the aggregation mechanism unchanged.

\begin{table}[h]
\centering
\small
\caption{GoT results on MuSR
(\texttt{gpt-oss-120b}, $n{=}100$).
Agg-$\Delta$ is the fraction of problems on which GoT aggregation changes the
beam answer; Agg-Correct is the fraction of changed answers that are correct.}
\label{tab:got}
\begin{tabular}{lccccc}
\toprule
\textbf{Method}
& \textbf{Acc.}
& $\bm{N_{\mathrm{eff}}}$
& \textbf{Escape\%}
& \textbf{Agg-$\Delta$\%}
& \textbf{Agg-Correct\%} \\
\midrule
GoT
& 0.570
& 5.35
& 48.3
& 17
& 64 \\
GoT + \textsc{BASIN}
& 0.600
& \textbf{5.69}
& \textbf{51.6}
& \textbf{18}
& \textbf{69} \\
GoT + \textsc{QA-BASIN}
& \textbf{0.640}
& 5.12
& 45.8
& 16
& 68 \\
\bottomrule
\end{tabular}
\end{table}

Flat \textsc{BASIN} improves GoT accuracy by $3$pp, from $57\%$ to $60\%$,
while increasing $N_{\mathrm{eff}}$ from $5.35$ to $5.69$.
\textsc{QA-BASIN} further improves accuracy to $64\%$, a $7$pp gain over
standard GoT, despite reducing $N_{\mathrm{eff}}$ to $5.12$ and the escape
rate to $45.8\%$. This again shows that improved reasoning does not require
maximizing exploration: quality-aware selection can retain stronger reasoning
regions while avoiding unproductive revisits.

The aggregation step changes the beam answer on 16--18\% of problems and
selects the correct answer in 64--69\% of those cases. The mechanisms therefore
act at complementary stages: basin-aware selection determines which strategies
survive search, while GoT aggregation combines the resulting trajectories
afterward. These results provide additional evidence that both the basin
mechanism and its quality-aware extension transfer beyond a single
tree-search controller.

\section{Sensitivity to the Quality Signal}
\label{app:qa_basin_verifier}

The quality-aware formulation in Eq.~\eqref{eq:qa_basin} assumes that the
signal used to estimate basin quality is informative. We therefore examine
both the quality of the underlying scoring signal and the sensitivity of
\textsc{QA-BASIN} to different choices of that signal.

\paragraph{Quality of heuristic scores.}
We first evaluate whether the heuristic scoring function used during search
reliably distinguishes correct from incorrect trajectories. Table~\ref{tab:verifier_auc}
reports its discriminative performance on MuSR with
\texttt{gpt-oss-120b}. The score AUC is close to chance for both GoT and
GoT+\textsc{BASIN} (0.510 and 0.516, respectively), indicating that the
heuristic provides little direct information about trajectory correctness.
Nevertheless, \textsc{BASIN} improves final accuracy from 57.0
and selection efficiency from 0.671 to 0.706. This suggests that the gain in
this setting does not arise from a strong verifier, but from changing which
reasoning strategies survive search.

\begin{table}[h]
\centering
\small
\caption{Heuristic scoring quality on MuSR
(\texttt{gpt-oss-120b}, $n{=}100$, six rounds).
AUC is computed over terminal trajectories.}
\label{tab:verifier_auc}
\begin{tabular}{lccccc}
\toprule
\textbf{Method}
& \textbf{Acc.}
& \textbf{Pass@$k$}
& \textbf{Sel. Eff.}
& \textbf{Score AUC}
& \textbf{Agg. changed} \\
\midrule
GoT
& 0.570
& 0.850
& 0.671
& 0.510
& 17\% \\
GoT + \textsc{BASIN}
& \textbf{0.600}
& 0.850
& \textbf{0.706}
& \textbf{0.516}
& 18\% \\
\bottomrule
\end{tabular}
\end{table}

\paragraph{Effect on quality-aware selection.}
A weak quality signal is more consequential for \textsc{QA-BASIN}, because
the signal directly controls the strength of the basin-revisit penalty.
Table~\ref{tab:qa_verifier} compares \textsc{QA-BASIN} using the heuristic
score with an LLM-based quality signal on MuSR with \texttt{gpt-4}. Using
the weak heuristic substantially reduces accuracy to 33.7
standard ToT (52.0
\texttt{gpt-4o-mini} as the quality signal yields 58.7
over flat \textsc{BASIN} and substantially outperforming standard ToT.

\begin{table}[h]
\centering
\small
\caption{Effect of quality-signal choice on \textsc{QA-BASIN} for MuSR
(\texttt{gpt-4}, $\lambda{=}0.5$, $n{=}300$).}
\label{tab:qa_verifier}
\begin{tabular}{llccc}
\toprule
\textbf{Method}
& \textbf{Quality signal}
& \textbf{Acc.}
& $n_{\mathrm{basins}}$
& $\bm{N_{\mathrm{eff}}}$ \\
\midrule
Standard ToT
& --
& 0.520
& 5.86
& 4.67 \\
\textsc{BASIN}
& --
& 0.583
& 7.39
& 5.72 \\
\textsc{QA-BASIN}
& Heuristic
& 0.337
& 3.81
& 3.36 \\
\textsc{QA-BASIN}
& LLM (\texttt{gpt-4o-mini})
& \textbf{0.587}
& 7.32
& 5.64 \\
\bottomrule
\end{tabular}
\end{table}

These results clarify the roles of the two formulations. Flat
\textsc{BASIN} does not require an estimate of basin quality and can therefore
be used when no reliable quality signal is available. \textsc{QA-BASIN}
provides a better exploration--exploitation mechanism when an informative
quality signal is available, because it can preserve promising basins while
continuing to penalize repeated visits to lower-quality ones. However, a poor
quality signal can incorrectly protect weak basins or suppress useful
exploration, as illustrated by the heuristic result above. We therefore view
\textsc{QA-BASIN} as the preferred formulation when a meaningful verifier or
quality estimate is available, rather than as uniformly superior independent
of signal quality.

\section{Basin Definition: NLI versus Embedding Similarity}
\label{app:basin_def}

For open-ended tasks such as MuSR, reasoning basins require an approximate
semantic equivalence relation. We use NLI-based clustering because direct
embedding similarity produces overly coarse partitions of the reasoning space.
Table~\ref{tab:basin_def} compares the two representations on a MuSR
diagnostic subset.

\begin{table}[h]
\centering
\small
\caption{Basin structure under SBERT cosine similarity and NLI-based semantic
clustering on MuSR. Very small $N_{\mathrm{eff}}$ indicates that most
trajectories are assigned to the same basin.}
\label{tab:basin_def}
\begin{tabular}{llcc}
\toprule
\textbf{Basin definition}
& \textbf{Model}
& \textbf{Mean basins}
& $\bm{N_{\mathrm{eff}}}$ \\
\midrule
SBERT cosine
& \texttt{gpt-4o-mini}
& 1.40
& 1.37 \\
SBERT cosine
& \texttt{gpt-oss-120b}
& 1.05
& 1.04 \\
\midrule
NLI semantic
& \texttt{gpt-oss-120b}
& \textbf{8.58}
& \textbf{6.85} \\
\bottomrule
\end{tabular}
\end{table}

SBERT cosine similarity merges most MuSR trajectories into one or two basins.
Different reasoning traces for the same problem reuse substantial narrative
context, so explanations supporting different hypotheses can remain close in
embedding space. A representation with $N_{\mathrm{eff}}$ near one provides
little useful strategy-level structure because nearly every candidate is
treated as belonging to the same region.

NLI-based clustering instead compares the propositional content of compact
extracted hypotheses. Paraphrases supporting the same answer and argument can
be grouped together, while incompatible hypotheses remain separate. This
produces a substantially richer partition and better matches the type of
strategy-level redundancy that \textsc{BASIN} is intended to track.

More generally, exact symbolic or structural keys are preferable when they are
available. Semantic clustering is a fallback for tasks without a deterministic
equivalence relation; the selection mechanism itself is agnostic to how basin
membership is constructed.

\section{Semantic Basin Sensitivity}
\label{app:semantic_sensitivity}

Semantic basin construction introduces learned components and clustering
choices. We therefore rerun the full search while varying the entailment
threshold, NLI model, and hypothesis extractor rather than merely reclustering
fixed trajectories post hoc. Table~\ref{tab:semantic_sensitivity} reports
results for MuSR with \texttt{gpt-4o-mini}; standard ToT achieves accuracy
$0.607$ and Pass@$k=0.857$.

\begin{table}[h]
\centering
\small
\caption{Sensitivity of \textsc{BASIN} to semantic basin construction on
MuSR.}
\label{tab:semantic_sensitivity}
\begin{tabular}{lcccc}
\toprule
\textbf{Basin definition}
& \textbf{Acc.}
& \textbf{Pass@$k$}
& $n_{\mathrm{basins}}$
& $\bm{N_{\mathrm{eff}}}$ \\
\midrule
Default ($\tau_e{=}0.45$, v3-small, default extractor)
& 0.620
& 0.833
& 6.84
& 5.28 \\
$\tau_e{=}0.30$
& \textbf{0.640}
& 0.780
& 3.78
& 3.17 \\
$\tau_e{=}0.60$
& \textbf{0.640}
& 0.760
& 6.28
& \textbf{5.78} \\
Alternative NLI model (v3-base)
& \textbf{0.640}
& 0.780
& 4.36
& 3.74 \\
Alternative extractor (\texttt{gpt-4})
& 0.620
& 0.740
& 4.74
& 4.09 \\
\bottomrule
\end{tabular}
\end{table}

Accuracy remains between $0.620$ and $0.640$ and exceeds standard ToT under
every tested semantic basin definition despite substantial changes in basin
counts and $N_{\mathrm{eff}}$. Pass@$k$ is more sensitive, particularly to the
choice of hypothesis extractor. Thus, the semantic representation affects the
detailed search trajectory, but the observed accuracy improvement is robust
across the tested configurations.

The contradiction ceiling has no observable effect over
$\tau_c\in\{0.20,0.30,0.40\}$ in these experiments, suggesting that the
entailment criterion already removes most incompatible pairs. We nevertheless
view semantic basin construction as a genuine source of modeling sensitivity
and report these results to make that dependence explicit.

\section{Redundancy Gap as a Routing Signal}
\label{app:gap_routing}

The redundancy gap is informative about search behavior, but it is not
sufficient by itself as a deployment rule. We test whether
$\Delta_\rho$, computed from standard-search traces, predicts whether
\textsc{BASIN} improves accuracy. We additionally measure
$\Delta N_{\mathrm{eff}}$, the change in effective basin count under
\textsc{BASIN}, where larger values indicate a stronger increase in
strategy-level exploration.

Following the rebuttal analysis, we define four regimes using the signs of
$\Delta_\rho$ and $\Delta N_{\mathrm{eff}}$, with a threshold of $0.3$ for
the latter:
\begin{itemize}
    \item \textbf{Explore-clear:}
    $\Delta N_{\mathrm{eff}}>0.3$ and $\Delta_\rho\leq0$;
    \item \textbf{Exploit-clear:}
    $\Delta N_{\mathrm{eff}}\leq0.3$ and $\Delta_\rho>0$;
    \item \textbf{Restructuring:}
    $\Delta N_{\mathrm{eff}}\leq0.3$ and $\Delta_\rho\leq0$;
    \item \textbf{Ambiguous:}
    $\Delta N_{\mathrm{eff}}>0.3$ and $\Delta_\rho>0$.
\end{itemize}

The aggregate rule selects \textsc{BASIN} for the entire experiment whenever
$\Delta_\rho\leq0$. The combined rule uses the same decision except in the
Ambiguous regime, where it additionally uses a per-problem search-effort signal
already available from the standard run: the number of explored tree nodes
before termination, \texttt{n\_nodes}. Problems at or below the
experiment-specific median are routed to \textsc{BASIN}; those above the
median remain under standard ToT.

\begin{table}[h]
\centering
\small
\caption{\textbf{Redundancy gap as a routing signal.}
$\Delta_\rho$ is computed from standard-search traces and
$\Delta N_{\mathrm{eff}}$ is the change in effective basin count under
\textsc{BASIN}. Acc.\ (routed) uses the combined per-problem routing rule in
the Ambiguous regime. Agg.\ and Comb.\ indicate whether the aggregate and
combined rules, respectively, select the empirically better fixed policy.}
\label{tab:gap_routing}
\resizebox{\textwidth}{!}{%
\begin{tabular}{lrrccclcc}
\toprule
\textbf{Experiment}
& $\bm{\Delta_\rho}$
& $\bm{\Delta N_{\mathrm{eff}}}$
& \textbf{Acc.\ (std)}
& \textbf{Acc.\ (\textsc{BASIN})}
& \textbf{Acc.\ (routed)}
& \textbf{Regime}
& \textbf{Agg.}
& \textbf{Comb.} \\
\midrule
Game24 / \texttt{gpt-4o-mini}
& $+0.037$
& $+0.638$
& 0.660
& 0.720
& \textbf{0.760}
& Ambiguous
& No
& Yes \\
Game24 / \texttt{gpt-oss}
& $+0.486$
& $-0.184$
& \textbf{0.380}
& 0.370
& \textbf{0.380}
& Exploit-clear
& Yes
& Yes \\
Game24 / \texttt{Qwen3-397B}
& $+0.230$
& $+0.935$
& 0.380
& \textbf{0.490}
& 0.480
& Ambiguous
& No
& Yes \\
Game24 / \texttt{Qwen3-27B}
& $+0.114$
& $+2.502$
& 0.430
& 0.650
& \textbf{0.700}
& Ambiguous
& No
& Yes \\
BBH / \texttt{gpt-oss}
& $-0.430$
& $+0.921$
& 0.520
& \textbf{0.620}
& \textbf{0.620}
& Explore-clear
& Yes
& Yes \\
BBH / \texttt{gpt-4o-mini}
& $-0.349$
& $+0.356$
& \textbf{0.740}
& 0.700
& 0.700
& Explore-clear
& No
& No \\
\bottomrule
\end{tabular}}
\end{table}

Using $\Delta_\rho$ alone selects the empirically better fixed policy in only
$2/6$ settings ($33.3\%$). Its main failure mode is the Ambiguous regime:
all three Ambiguous experiments have $\Delta_\rho>0$, which would favor
standard search under the aggregate rule, yet \textsc{BASIN} improves accuracy
in all three. Incorporating the already-available \texttt{n\_nodes} signal
raises the routing decision accuracy to $5/6$ settings ($83.3\%$).

More importantly, per-problem routing improves over both globally fixed
policies in two of the three Ambiguous settings. On
Game24/\texttt{gpt-4o-mini}, routed accuracy reaches $0.760$, compared with
$0.660$ for standard ToT and $0.720$ for \textsc{BASIN}. On
Game24/\texttt{Qwen3-27B}, routing reaches $0.700$, compared with $0.430$ and
$0.650$, respectively. In the remaining Ambiguous setting, routing remains
close to the better fixed policy ($0.480$ vs.\ $0.490$).

These results reinforce the interpretation used in the main paper:
the redundancy gap is a useful diagnostic of search structure, but it is
insufficient as a standalone criterion for deciding when to increase
exploration. Combining redundancy information with inexpensive problem-level
search-state signals provides a more promising basis for adaptive
structure-aware search.

\section{Case Study: MuSR Murder Mystery:
\texttt{murder\_mysteries\_185}}
\label{app:example}

This MuSR example illustrates reasoning basin collapse in a semantic setting.
The story concerns the death of Wilhelmina by crossbow. Two salient suspects
are Isabelle and Nicole. Isabelle is a yoga instructor and member of an
archery club who was present in the kitchen during the murder. Nicole owns an
authentic medieval crossbow, remained at the crime scene throughout the day,
and is associated with a pattern of suspicious deaths among acquaintances. The
correct answer is \textbf{Nicole}.

Under standard ToT, the search predicts \textbf{Isabelle}. Although the
surface forms of the generated explanations differ, most terminal hypotheses
reuse the same core strategy: Isabelle had crossbow-related skill and was
present at the scene. Representative hypotheses include:

\begin{quote}\small\itshape
``Isabelle is most likely the murderer because she is a skilled crossbow
practitioner through her archery-club membership and was physically present in
the kitchen at the time of Wilhelmina's death.''
\end{quote}

\begin{quote}\small\itshape
``The most probable culprit is Isabelle---her confirmed crossbow expertise and
attendance at the yoga session held in the victim's kitchen at the time of the
killing place her squarely at the scene.''
\end{quote}

\begin{quote}\small\itshape
``Isabelle committed the murder: she practises crossbow shooting regularly and
her own account places her in Wilhelmina's kitchen during the window of the
crime.''
\end{quote}

\noindent
These trajectories differ lexically but are equivalent at the strategy level:
they predict the same answer and rely on the same central hypothesis
(\emph{crossbow skill} plus \emph{presence at the scene}). They are therefore
assigned to the same semantic reasoning basin. Repeated selection from this
basin causes the search budget to elaborate the same explanation rather than
testing materially different alternatives.

\textsc{BASIN} reduces the relative score of further revisits to the
Isabelle-centered basin, allowing the search to retain a distinct
Nicole-centered explanation:

\begin{quote}\small\itshape
``Nicole is the most likely murderer: she owns a genuine medieval crossbow
displayed in her home, the victim was killed in Nicole's own kitchen during a
visit Nicole hosted, and multiple people in Nicole's social circle have died
under similarly mysterious circumstances.''
\end{quote}

\noindent
This trajectory belongs to a different semantic basin: it predicts
\emph{Nicole} and relies on a different explanatory strategy combining
\emph{weapon ownership}, \emph{crime-scene ownership}, and a
\emph{pattern of suspicious deaths}. Preserving this alternative changes the
composition of the final candidate set and allows the correct answer to be
selected.

Together with the symbolic case study in the main paper, this example
illustrates the common mechanism underlying reasoning basin collapse. The
surface manifestation differs across domains, but in both cases search spends
multiple selections on states that instantiate the same underlying strategy.
\textsc{BASIN} uses this structure to discourage redundant revisits and
reallocate inference budget toward underexplored alternatives.

\section{Prompt Templates}
\label{app:prompts}

All generation prompts are identical between standard search and
\textsc{BASIN}; the methods differ only in the selection rule. MuSR additionally
uses the same hypothesis-extraction and semantic-clustering pipeline for all
compared search conditions. We list the principal prompts below.

\subsection*{Game of 24 --- Step Proposal}

At each search depth, the model receives the current remaining numbers and
proposes up to five candidate next steps. The system instruction and few-shot
examples are fixed; only the final \texttt{Input:} line changes.

\begin{tcolorbox}[title=Game of 24 proposal prompt,
fontupper=\small\ttfamily,
breakable, left=4pt, right=4pt]
\textbf{[System]} You are a Game of 24 expert. At each step, you are given
the CURRENT remaining numbers. Choose exactly two of them, apply one
arithmetic operation (+, -, *, /), and list possible next steps. Format
each step as: A op B = C (remaining: X Y Z) where X Y Z are the numbers
left after replacing A and B with C. List up to 5 diverse candidate steps.

\medskip
Input: 4 4 6 8\
Possible next steps:\
4 + 8 = 12 (remaining: 4 6 12)\
6 - 4 = 2 (remaining: 2 4 8)\
4 * 6 = 24 (remaining: 4 8)\
4 * 8 = 32 (remaining: 4 6)\
6 + 8 = 14 (remaining: 4 4 14)

\medskip
Input: 2 9 10 12\
Possible next steps:\
12 * 2 = 24 (remaining: 9 10)\
2 + 12 = 14 (remaining: 9 10 14)\
10 - 2 = 8 (remaining: 8 9 12)\
10 - 9 = 1 (remaining: 1 2 12)\
9 + 12 = 21 (remaining: 2 10 21)

\medskip
\textit{[additional few-shot examples omitted for brevity]}

\medskip
Input: \textit{{remaining numbers}}\
Possible next steps:
\end{tcolorbox}

\subsection*{MuSR --- Reasoning Generation}

Each generation round produces one candidate reasoning trajectory. The system
prompt is shared across rounds and search methods.

\begin{tcolorbox}[title=MuSR generation prompt,
fontupper=\small\ttfamily,
breakable, left=4pt, right=4pt]
\textbf{[System]} You are a careful reasoning assistant. Always end your
response with `Answer: X' on its own line, where X is a single uppercase
letter (A, B, C, \ldots). Never skip the Answer line.

\medskip
\textbf{[User]} \textit{{full MuSR problem text and answer choices}}
\end{tcolorbox}

\subsection*{MuSR --- Structured State Extractor}

After each reasoning-generation call, a separate
\texttt{gpt-4o-mini} call extracts a compact structured representation from
the raw trace. The \texttt{main\_hypothesis} field is used as the semantic
representation for NLI-based basin construction; the remaining fields support
downstream analysis and selection.

\begin{tcolorbox}[title=MuSR extractor prompt,
fontupper=\small\ttfamily,
breakable, left=4pt, right=4pt]
You are given a reasoning trace that answers a multiple-choice question.
Extract the following information as a JSON object with EXACTLY these five
keys:

\medskip
\quad `final\_answer''\ \ \ \ : the answer letter chosen (A / B / C / D / E)\\
\quad `main\_hypothesis'' : ONE sentence stating the single most important\
\quad\quad\quad\quad\quad\quad\quad\quad\quad\quad argument for that answer\
\quad `key\_evidence''\ \ \ : a JSON list of 2--3 short strings citing specific\\
\quad\quad\quad\quad\quad\quad\quad\quad\quad\quad facts from the story\\
\quad `eliminated\_option'': ONE sentence naming which option was ruled out\
\quad\quad\quad\quad\quad\quad\quad\quad\quad\quad and the decisive reason\
\quad ``reasoning\_summary'': ONE sentence describing the overall reasoning\
\quad\quad\quad\quad\quad\quad\quad\quad\quad\quad strategy used

\medskip
Return ONLY the JSON object --- no markdown fences, no explanation, nothing
else.

\medskip
Reasoning trace:\
`{}`{}`\\
\textit{\{trace\}}\\
`{}`{}`
\end{tcolorbox}

\section{Compute Resources}
\label{app:compute}

\textbf{LLM inference.}
LLM inference is performed through external model APIs, so the experiments do
not require local GPU inference. Experiments are parallelized across problems
using multi-threaded API calls. The broader camera-ready evaluation includes
the model families reported in the main tables, while the original MuSR and
Game-of-24 experiments use \texttt{gpt-4o-mini},
\texttt{gpt-4}, \texttt{gpt-oss-120b}, and
\texttt{Qwen3-27B}.

\textbf{Semantic representation cost.}
For MuSR, both standard ToT and basin-aware search use the same semantic
representation pipeline. Each problem uses 18 reasoning-generation calls and
18 hypothesis-extraction calls, for 36 LLM calls in total. Thus, hypothesis
extraction adds 100
identical across the compared MuSR search conditions. Consequently, the MuSR
experiments isolate the effect of the selection rule conditional on using the
same extraction and semantic-clustering machinery.

The local neural component is the NLI model used for semantic basin assignment,
\texttt{cross-encoder/nli-deberta-v3-small}, which runs on CPU. A
bidirectional pairwise comparison takes approximately $11.8$,ms when batched
at 32. Clustering roughly 18 states requires approximately 4--11,s per
problem, plus a one-time 2.6,s model-loading cost per worker. The current
implementation recomputes same-answer pairwise scores rather than updating the
clusters incrementally, so these timings should be viewed as those of an
unoptimized implementation.

For tasks with deterministic structural basin definitions, no hypothesis
extraction or NLI computation is required for basin assignment. The additional
search-side computation is limited to constructing the basin key, maintaining
visit statistics, and modifying the candidate-selection score.

\textbf{Hardware.}
Local experiments were run on a MacBook Pro with an Apple M2 8-core CPU and
16,GB unified memory. No local GPU is required for the reported basin
tracking or semantic-clustering computations.

\textbf{API usage.}
The experiments require substantially more API inference than a single
generation baseline because inference-time search evaluates multiple reasoning
trajectories. On MuSR, the semantic representation pipeline further doubles
the number of LLM calls relative to generation alone, as described above.
We therefore do not characterize semantic basin construction as computationally
negligible; its cost is a limitation of the current open-ended implementation.


\end{document}